\documentclass[letterpaper, 10 pt, conference]{ieeeconf}  
\usepackage{amsmath}
\usepackage{graphicx}
\usepackage{multirow}
\usepackage{float}
\usepackage{pbalance}
\usepackage{balance}
\usepackage[linesnumbered,ruled,vlined]{algorithm2e}
\IEEEoverridecommandlockouts                              

\title{\LARGE \bf
GazeRace: Revolutionizing Remote Piloting with Eye-Gaze Control 
}

\author{Issatay Tokmurziyev\textsuperscript{*}, Valerii Serpiva\textsuperscript{*}, Aleksey Fedoseev, Miguel Altamirano Cabrera, Dzmitry Tsetserukou
\thanks{The authors are with the Intelligent Space Robotics Laboratory, Skolkovo Institute of Science and Technology Moscow, Bolshoy Boulevard 30, bld. 1, 121205, Moscow, Russia. 
Email: {\tt\small \{Issatay.Tokmurziyev, Valerii.Serpiva, Aleksey.Fedoseev, Miguel.Altamirano, D.Tsetserukou\}@skoltech.ru\ 
}
}
\thanks{*These authors contributed equally to this work.}
}

\begin{document}

\maketitle
\thispagestyle{empty}
\pagestyle{empty}

\begin{abstract}



This paper presents GazeRace, a novel system that leverages eye-tracking technology for intuitive drone control. Using the MediaPipe library, the system translates eye movements into precise drone commands, enabling effective remote piloting. In testing, GazeRace demonstrated an 18\% reduction in drone trajectory length while maintaining competitive speed with traditional controls. The results suggest that this approach enhances control accuracy and reduces user frustration, offering a significant advancement in the field of human-computer interaction and drone navigation.
\end{abstract}
\noindent



\section{Introduction}

The advancements in drone technology have significantly broadened the scope of their applications. Initially utilized for general purposes, drones can now fulfil diverse functions across various sectors. Their usefulness ranges from recreational activities to pivotal roles in surveillance, disaster response, logistics support, and agricultural management \cite{droneimpact}. Typically, drones are remotely operated by a human operator, requiring manual control from the ground through a remote controller or ground-based controls, which mandate the operator to maintain visual contact and manually input each command. Although this method is useful, its manual nature requires the operator to be highly cognitively and physically engaged; therefore, it can be difficult, especially in challenging environments during extended operations.

\begin{figure}[t]
    \centering
    \includegraphics[width=1\linewidth]{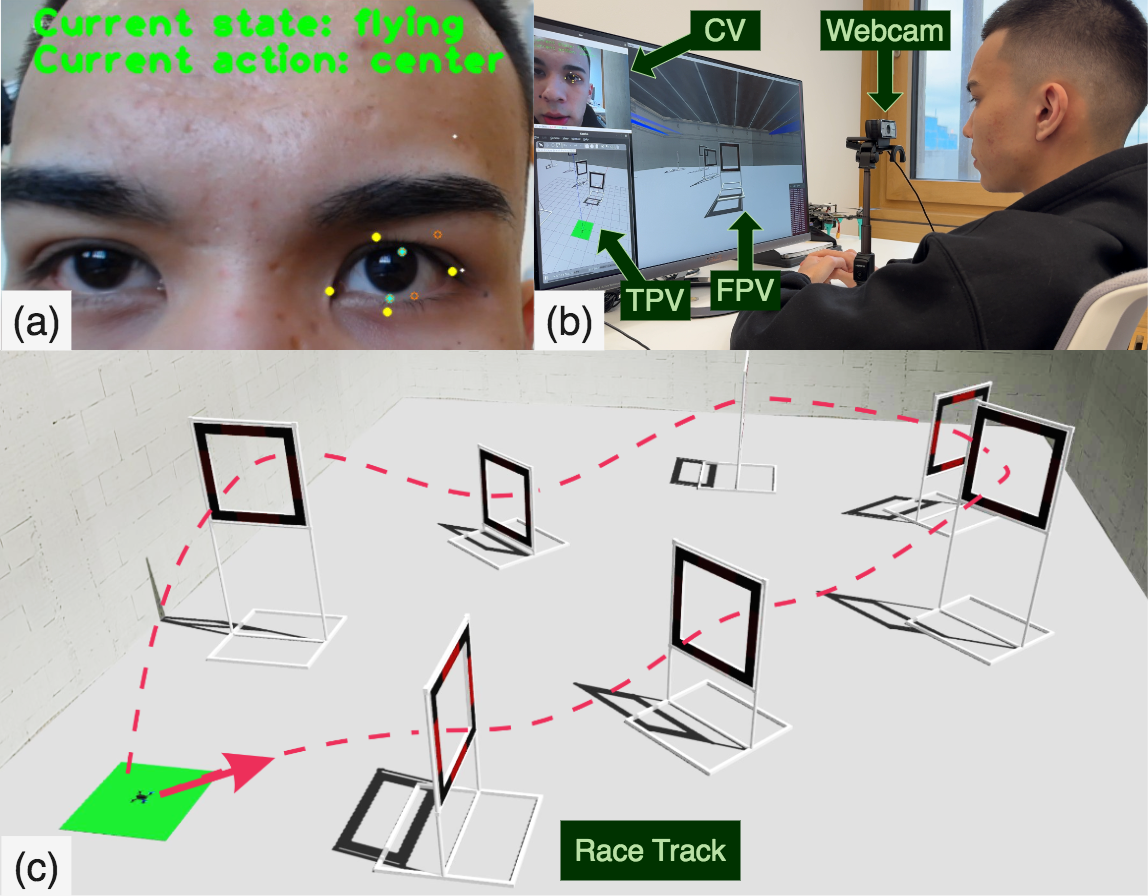}
    \caption{GazeRace interface for guiding racing drone: (a) MediaPipe interface with predefined right eye landmarks. (b) Experimental setup for the evaluation of eye control interface. (c) Drone race track in a Gazebo simulation environment.}
    \label{fig:tiser}
\end{figure}

Innovations in human-computer interfaces promise to enhance drone control systems by providing more intuitive and flexible designs capable of significantly reducing the cognitive and physical burdens on the user \cite{mueller2020next}. For example, a revolutionary technology within this domain is eye-gaze control \cite{klaib2021eye}. This allows the user's gaze direction to serve as an input mechanism for controlling drone movements. It harnesses the innate human ability to rapidly and accurately shift attention by moving the eyes. Hence, it offers a powerful tool for remote control scenarios.

This paper explores one possible domain of employment for eye-tracking technologies in drone control, proposing the integration of sophisticated eye-tracking systems with advanced Unmanned Aerial Vehicle (UAV) control frameworks. The core of this approach is the application of MediaPipe, utilizing state-of-the-art Computer Vision (CV) algorithms to precisely localize and track the user's iris position in real-time. The proposed methodology involves translating gaze directions into drone commands via a heads-up, hands-free system to facilitate intuitive UAV navigation. This has the potential to revolutionize remote piloting and holds promise for significant improvements in this area.

Hands-free eye-tracking control for drones not only enhances user convenience but also improves operational safety and accessibility. This technology has the potential to provide new opportunities for people with motor disabilities or paralysis to operate drones, eliminating the need for traditional physical controls. Rigorous testing and experimental validation demonstrate the capabilities and benefits of integrating eye-tracking into UAV operations, setting the stage for its widespread application in various aspects of the drone industry.

\section{Related Work}

\subsection{Human-Robot Interaction}

Human-Robot Interaction (HRI) is rapidly evolving, particularly through innovative methods allowing intuitive communication between humans and robots. Many research papers are exploring ways to intuitively control UAVs and drones \cite{tezza2019state}. Research on drone and robot interaction with humans in shared environments has demonstrated their potential for real-world applications. For example, studies have shown that human-robot collaboration is effective in delivering medical supplies \cite{hri_med} and in firefighting operations \cite{hri_fire}. Currently, there are challenges in implementing human-machine interaction. The control interfaces have various limitations, and the human factor cannot be excluded \cite{Challenges}.

Advancements in HRI are enhancing the intuitiveness and efficiency of communication between humans and robots. The integration of natural user interfaces, like voice commands and eye-tracking, along with multimodal interaction systems, is making robot operations more accessible and context-aware. This integration plays a crucial role in enabling robots to predict and adapt to human intentions, improving workflow and safety in dynamic environments.

\subsection{Non-Conventional Drone Control Approaches}

Hand gesture interfaces have been extensively investigated for the motion control of the drone without any handheld device. For example, the gesture input interface is suggested by Suresh et al. \cite{Suresh_2019}, where arm gestures and motions are recorded by a wearable armband and applied to control a swarm formation. A haptic glove with finger tracking for human-swarm interaction with an impedance-controlled swarm of drones was introduced by Tsykunov et al. \cite{Tsykunov_2019}. Fedoseev et al. \cite{Fedoseev_2022} proposed a concept of haptic drones that were controlled by human hand motion tracking from a virtual reality environment. A wearable interface with multi-finger motion detection and capacitive sensing to control a single aerial drone is proposed by Montebaur et al. \cite{Montebaur_2020}. The interface is intended for use with special operations forces, where it is important to maintain a high mobility of the user. Srivastava et al. \cite{Srivastava_2020} implemented a gesture-detecting bracelet for intuitive control over several drones. Serpiva et al. \cite{Serpiva_2021} proposed a gesture input interface based on the MediaPipe framework for direct teleoperation and trajectory drawing of a drone swarm.

Aside from hand tracking, other body tracking concepts were suggested as input interfaces for drone control. For example, Rognon et al. \cite{Rognon_2018} proposed a full-body exoskeleton for the precise control of a drone. The developed system allowed to track the position and orientation of the human body with high accuracy, however, the wearable system was bulky for convenient usage. The tracking of upper body motion with a remote camera was proposed by Miehlbradt et al. \cite{Miehlbradt_2018} providing a more convenient interface with drone control by human torso and arm motion. Baza et al. \cite{Baza_2022} applied full-body tracking with a camera to control a human avatar composed of a swarm of drones. While allowing to control several copters through a single input, this approach is less applicable for a racing drone that is required to move at a high speed. Additionally, brain-computer interfaces (BCIs) are being developed to control drones using EEG signals, pushing the boundaries of how humans might interact with robots without traditional physical controls \cite{9154926}. These advances collectively point towards a future where HRI is increasingly seamless, intuitive, and inclusive.

One study highlights a system where the human gaze directs a Micro Aerial Vehicle (MAV), integrating eye-tracking glasses to spatially task the robot, emphasizing non-invasive interaction techniques that could also assist individuals with mobility impairments \cite{yuan2019human}. Another approach, GPA-Teleoperation, enhances this interaction by using gaze to infer operator intentions for drone teleoperation, effectively making the system accessible to non-professional users \cite{DBLP:journals/corr/abs-2109-04907}. Klaib et al. \cite{kim2014quadcopter} provides an overview of eye-tracking methods, focusing on modern techniques such as Machine Learning (ML), the Internet of Things (IoT), and cloud computing. The study referenced in \cite{inproceedings} shares similarities with ours, as both involve participants completing a navigational task using different control methods. However, while the study assesses both gaze-based and keyboard-based controls, our research focuses solely on standalone gaze control.




\section{Drone Control Scenario}

\subsection{System Overview}
The developed GazeRace interface consists of an eye-tracking system based on real-time iris and eyelid tracking using CV, an algorithm for generating commands and actions for a drone, as well as a drone control system. It employs a Logitech C930e camera to capture users' eye movements. These eye movements are processed in real-time by the MediaPipe library, which uses a combination of Convolutional Neural Networks (CNNs) and Recurrent Neural Networks (RNNs) to extract features from face images. The neural network within MediaPipe is pre-trained to accurately detect and track key facial landmarks, particularly the eyes. This detection is crucial for the eye-tracking mechanism, as it enables the system to continuously monitor and interpret the user's eye movements. The identified landmarks are then used to calculate the position of the iris relative to other defined points around the eye to determine gaze direction or eye action. These actions are translated into commands for the SITL (Software In The Loop) drone within Gazebo simulation environments. The SITL ArduPilot Gazebo plugin simulates drone actions using the same control system via the MAVROS interface. This facilitates a thorough simulation of the drone's control procedures. All modules function and communicate via ROS nodes. Figure \ref{fig:Block diagram for two control methods} presents the system's block diagram.

The system was tested in a simulated environment within the Gazebo framework, where users navigated a drone through a predefined racing track. Experimental results demonstrated that users effectively controlled the drone using eye movements, significantly reducing the drone's trajectory length. 


\subsection{CV-based Iris Direction Recognition} 
\begin{figure}
    \centering
    \includegraphics[width=0.75\linewidth]{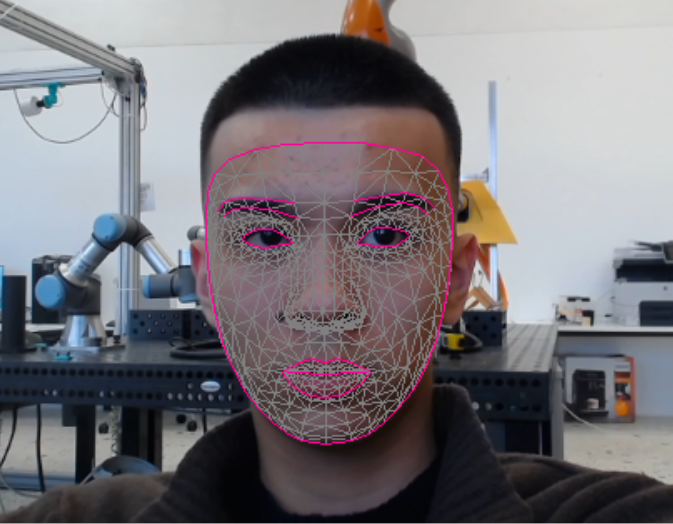}
    \caption{Face mesh reconstruction with MediaPipe framework utilized for eye-tracking based drone control of GazeRace.}
    \label{fig:face}
\end{figure}

The MediaPipe library provides the solution for exceptionally accurate face mesh reconstruction. To achieve precise tracking of the human operator, we pinpoint 468 distinct facial landmarks (see Fig. \ref{fig:face}), including specific markers around the iris, within each frame of the human’s face. These landmarks are then used to determine the position of the iris within the eye socket. 

\begin{figure*}
    \centering
    \includegraphics[width=\textwidth]{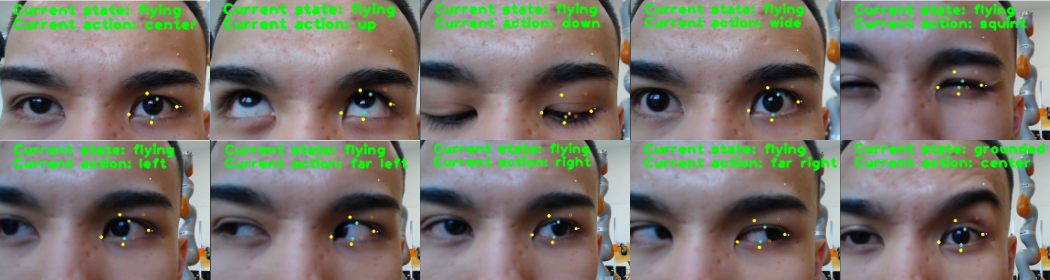}
    \caption{Ten actions ( up, down, left, right, far-left, far-right, wide, squint, center, and one eyebrow action) utilized for a low-delay drone motion control.}
    \label{fig:Ten}
\end{figure*}

First, the user should pass the calibration part. Upon launching the system, the camera initializes and the user is presented with instructions on the screen. These instructions guide the user to focus on specific points displayed on the screen and perform certain actions, such as widening their eyes, squinting, or looking in different directions. This process allows the system to accurately map the user's gaze and eyelid movements, ensuring that the subsequent control commands generated by the eye-tracking interface are precise and responsive to the user's inputs. The calibration ensures that each user's unique eye movement patterns are effectively captured and interpreted by the system, which is crucial for accurate and intuitive drone control. The process is described in Algorithm 1. Overall, the proposed control method utilizes nine actions from the right eye only, which include: up, down, left, right, far-left, far-right, wide, squint, center, and one eyebrow action named raise (Fig. \ref{fig:Ten}). The applied actions are then transformed to control the drone’s pitch, roll, yaw and throttle rate: 
\begin{figure}
    \centering
    \includegraphics[width=0.9\linewidth]{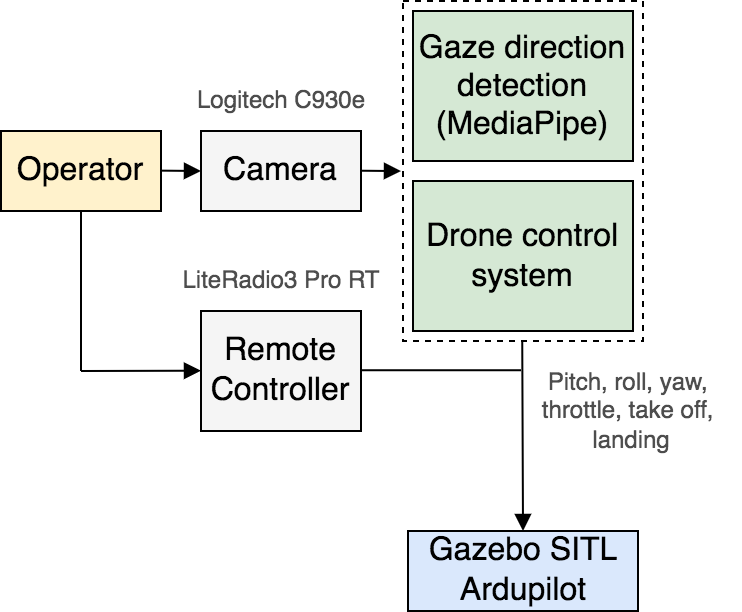}
    \caption{The pipeline of the drone mixed control approach based on the remote controller and the developed eye-tracking interface.}
    \label{fig:Block diagram for two control methods}
\end{figure}
\begin{itemize}
    \item \textbf{Raise} action handles drone arming and taking off. If triggered again, it handles drone landing and disarms the drone.
    \item \textbf{Wide/Squint} actions move the drone forward and backwards (Pitch).
    \item \textbf{Left/Right} actions move the drone left and right (Roll).
    \item \textbf{Far-left/Far-right} actions rotate the drone along its vertical axis (Yaw).
    \item \textbf{Up/Down} actions move the drone relatively (Throttle).
    \item \textbf{Center} action is a neutral one, once activated, the drone does not do anything.
\end{itemize}

\begin{algorithm}
\KwIn{Virtual Environment (Gazebo), Eye-Gaze Data, ROS Framework}
\KwOut{Control commands to the drone via ROS}
\textbf{Initialize} the drone in the Gazebo virtual environment\;
\textbf{Start} MediaPipe eye-gaze-tracking module\;

\For{each user}{
    \textbf{Calibration Stage:}\;
    \For{each of the 10 actions}{
        Calculate the eye-gaze ratio using formula (1)
    }
}

\While{True}{
    \textbf{Read} eye-gaze action\;
    \textbf{Send} action data to ROS\;
    \uIf{action is \textbf{Raise}}{
        \textbf{Prioritize} this action\;
        \uIf{drone is disarmed}{
            Arm the drone and take off\;
        }
        \Else{
            Land the drone and disarm\;
        }
    }
    \Else{
        \textbf{Translate} the eye-gaze action into a drone command:
        \uIf{action is \textbf{Wide}}{
            Move drone forward (Pitch)\;
        }
        \uIf{action is \textbf{Squint}}{
            Move drone backward (Pitch)\;
        }
        \uIf{action is \textbf{Left}}{
            Move drone left (Roll)\;
        }
        \uIf{action is \textbf{Right}}{
            Move drone right (Roll)\;
        }
        \uIf{action is \textbf{Far-left}}{
            Rotate drone left (Yaw)\;
        }
        \uIf{action is \textbf{Far-right}}{
            Rotate drone right (Yaw)\;
        }
        \uIf{action is \textbf{Up}}{
            Move drone up (Throttle)\;
        }
        \uIf{action is \textbf{Down}}{
            Move drone down (Throttle)\;
        }
        \uIf{action is \textbf{Center}}{
            Keep the drone in a neutral position\;
        }
    }
    
}
\caption{Drone Control Algorithm using MediaPipe Eye-Gaze-Tracking and ROS}
\end{algorithm}

The process of determining the particular action involves the calculation of ratios between the center of the iris and the predefined points within the eye socket. This ratio is calculated as:
\begin{equation}
R = \frac{\sqrt{(x_{\text{e1}} - x_{\text{c}})^2 + (y_{\text{e1}} - y_{\text{c}})^2}}{\sqrt{(x_{\text{e1}} - x_{\text{e2}})^2 + (y_{\text{e1}} - y_{\text{e2}})^2}} , 
\end{equation}
where $x_{\text{e1}}$, $y_{\text{e1}}$, $x_{\text{e2}}$, $y_{\text{e2}}$ are the coordinates of the first and second edge points, and $x_{\text{c}}$, $y_{\text{c}}$ are the coordinates of the center point. 
Then each of the actions is assigned its ratio values. The web camera is required to be at a defined distance from a user to be able to recognize their face and track the positions of all key points. 

\section{Experimental Evaluation}

\subsection{Research Methodology}

In this study, we conducted a comparative assessment of an eye control interface in comparison to a baseline remote controller. Ten participants (two female, mean age 24.7, SD=5.6) were invited to perform a drone race within the Gazebo simulation environment (Fig. \ref{fig:tiser}(c)). The participants were informed about the experimental procedure and agreed to the consent form. Users applied two different interfaces to navigate the drone's flight on a racing track. The user positioned himself in front of the camera (Fig. \ref{fig:tiser}(b)), causing it to capture his eyes (Fig. \ref{fig:tiser}(a)). At the same time, he was able to see the view from the drone camera, which enabled him to understand how to direct the drone's movements. The trajectories of their flights were recorded and analyzed. Half of the participants initially piloted the drone using a remote controller LiteRadio3 Pro RT, then transitioned to utilizing the developed eye control interface. The other $50\%$ tested the interfaces in reverse order. Short breaks were incorporated between attempts and during transitions between control methods to ensure fairness. Each participant was tasked with completing a race course comprising seven square gates with a size of 1.4 m. Before the race, a 30-minute familiarization period was provided for participants to adapt to both control methods. Before the eye-tracking-based control of the drone, each participant underwent calibration with the camera to ensure accurate detection of their eye motion.

Subsequently, each participant executed five flights with each interface. The best three attempts were evaluated. The race track was designed to challenge participants' proficiency in utilizing all control inputs necessary for drone operation (including yaw, pitch, roll, and throttle), whether through eye gaze control or baseline remote control interface, i.e. Logitech Wireless Gamepad F710. Given the varying levels of expertise among participants, the drone in the simulation was set to operate in the “position hold" mode throughout the experiment. Following the trials, each participant completed a comprehensive user survey consisting of two parts: the unweighted NASA-TLX (evaluated on a 100-point scale derived from a 21-point scale) and the UEQ questionnaires. These surveys aimed to assess both the pragmatic and hedonic quality of the eye-tracking interface, as well as its overall performance.

\subsection{Experimental Results}

Table \ref{tab:traj_results} presented the results of evaluation flight trajectories, and the results of the user questionnaire are presented in Table \ref{tab:survey}.

\begin{figure}[t]
\centerline{\includegraphics[width=0.5\textwidth]{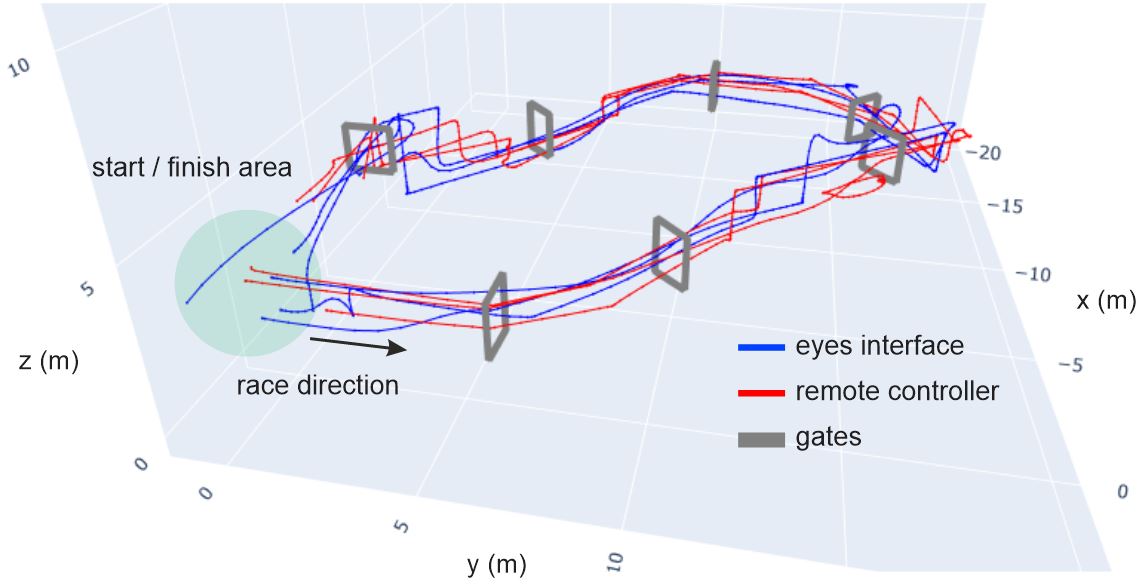}}
\caption{Set of recorded trajectories by the remote controller (red lines), and by eye-tracking interface (blue lines) with shortest path and flight time.}
\label{fig:traj_race}
\end{figure}

\begin{table}[b]
\caption{Experimental Results of Recorded Flight Trajectories by Eye-Tracking Interface and Remote Controller.} 
\label{tab:traj_results}
\centering
    \begin{tabular}{ |l|l|c|c| } 
        \hline
        \textbf{Participants  } & \textbf{Metrics} & \begin{tabular}[c]{@{}c@{}} \textbf{Eye-Tracking}\\ \textbf{Interface}\end{tabular} & \begin{tabular}[c]{@{}c@{}}\textbf{Remote}\\ \textbf{Controller}\end{tabular} \\
        \hline
        \multirow{4}{4em}{Overall}
        & Time, s &  70.01 & \textbf{67.5} \\ 
        & Path length, m & {{\textbf{73.44}}} & 89.29 \\ 
        & Average velocity, m/s & 1.08 & {{\textbf{1.39}}} \\ 
        & Maximal velocity, m/s & 3.25 & {{\textbf{6.52}}} \\ 
        \hline
        \multirow{3}{4em}{Best Result}
        & Time, s & 44.78 & \textbf{36.60} \\ 
        & Path length, m & \textbf{58.32} & 60.78 \\      
        & Average velocity, m/s & 1.56 & {{\textbf{2.35}}} \\ 
        \hline
    \end{tabular}
\end{table}

Upon evaluation, it was observed that the users were able to decrease the length of the drone trajectory on average by 18\% (73.44 m against 89.29 m) while successfully navigating between the racing gates with the developed eye-tracking interface. On the other hand, the trajectories achieved with the baseline remote controller were achieved within a 3.5\% lower time for an average participant (mean 67.5 s) compared to those achieved using the eye-tracking interface (mean 70.01 s). This small difference might be potentially caused by the lower confidence of users who initially used the eye tracker for the first time. The best time for GazeRace was 44.78 s with the shortest flight path of 58.32 m. The best time for remote control was 36.6 s with the shortest flight path of 60.73 m. Additionally, the trajectories appeared smoother when controlled via an eye-tracking interface rather than by a remote controller, as shown in Fig. \ref{fig:traj_race}. Also, the results show that 4 out of 10 participants completed the race faster using the GazeRace compared to the remote controller, on average 25.9\% faster.

\begin{table}[t]
    \caption{User Score of the Developed Interface with Unweighted NASA-TLX and UEQ Scores, $p$ $\textless$ 0.05 Highlighted.}
    \label{tab:exp_results}
\centering
    \begin{tabular}{ |c|c|c|c| } 
        \hline
        \textbf{Score} & \begin{tabular}[c]{@{}c@{}}\textbf{Remote}\\ \textbf{Controller}\end{tabular} & \begin{tabular}[c]{@{}c@{}}\textbf{Eye-Tracking}\\ \textbf{Interface}\end{tabular}  & \begin{tabular}[c]{@{}c@{}}\textbf{Wilcoxon,} \\ \textbf{alpha of 0.05}\end{tabular} \\
        \hline       
        \multicolumn{1}{|c}{NASA-TLX}& \multicolumn{3}{|c|}{} \\        
        \hline        
        \multirow[t]{7}{*}{Mental Demand} & \textbf{39.0} & 52.5 &\textbf{V=2.5, $p$=0.017} \\
        Physical Demand &\textbf{32.5} & 45.0 & \textbf{V=0.0, $p$=0.011} \\ 
        Temporal Demand & \textbf{48.5} & 60.5 & V=10.0, $p$=0.083\\ 
        Performance & 63.0 & \textbf{34.0} & \textbf{V=0.0, $p$=0.007}\\
        Effort & 49.0 & \textbf{30.5} &\textbf{V=1.0, $p$=0.003}\\
        Frustration & 46.5 &\textbf{22.5} &\textbf{V=0.0, $p$=0.002} \\
        Overall & 46.4 & \textbf{40.8} &\textbf{V=6.0, $p$=0.027} \\
        \hline    
        \multicolumn{1}{|c}{UEQ}& \multicolumn{3}{|c|}{} \\        
        \hline        
        \multirow[t]{7}{*} 
        {Attractiveness} & 0.33 & \textbf{1.03} & \textbf{V=15.5, $p$=0.023} \\
        Perspicuity & -0.72 & \textbf{-0.55} & \textbf{V=7.0, $p$=0.037} \\ 
        Efficiency &  \textbf{1.25} & 0.13 & \textbf{V=5.0, $p$=0.037} \\ 
        Dependability & 0.45 & \textbf{0.63} & V=23.0, $p$=0.69 \\
        Stimulation & 0.33 & \textbf{ 1.45} & \textbf{V=0.0, $p$=0.007}\\
        Novelty & -1.22 & \textbf{1.85} & \textbf{V=0.0, $p$=0.002} \\
        Hedonic Quality & -0.45 &\textbf{1.65} & \textbf{V=0.0, $p$=0.002} \\
        Pragmatic Quality & 0.32 & \textbf{0.56} & V=17.0, $p$=0.32 \\ 
        
        \hline
    \end{tabular}
    \label{tab:survey}
\end{table}

\begin{figure}[t]
\centerline{\includegraphics[width=0.5\textwidth]{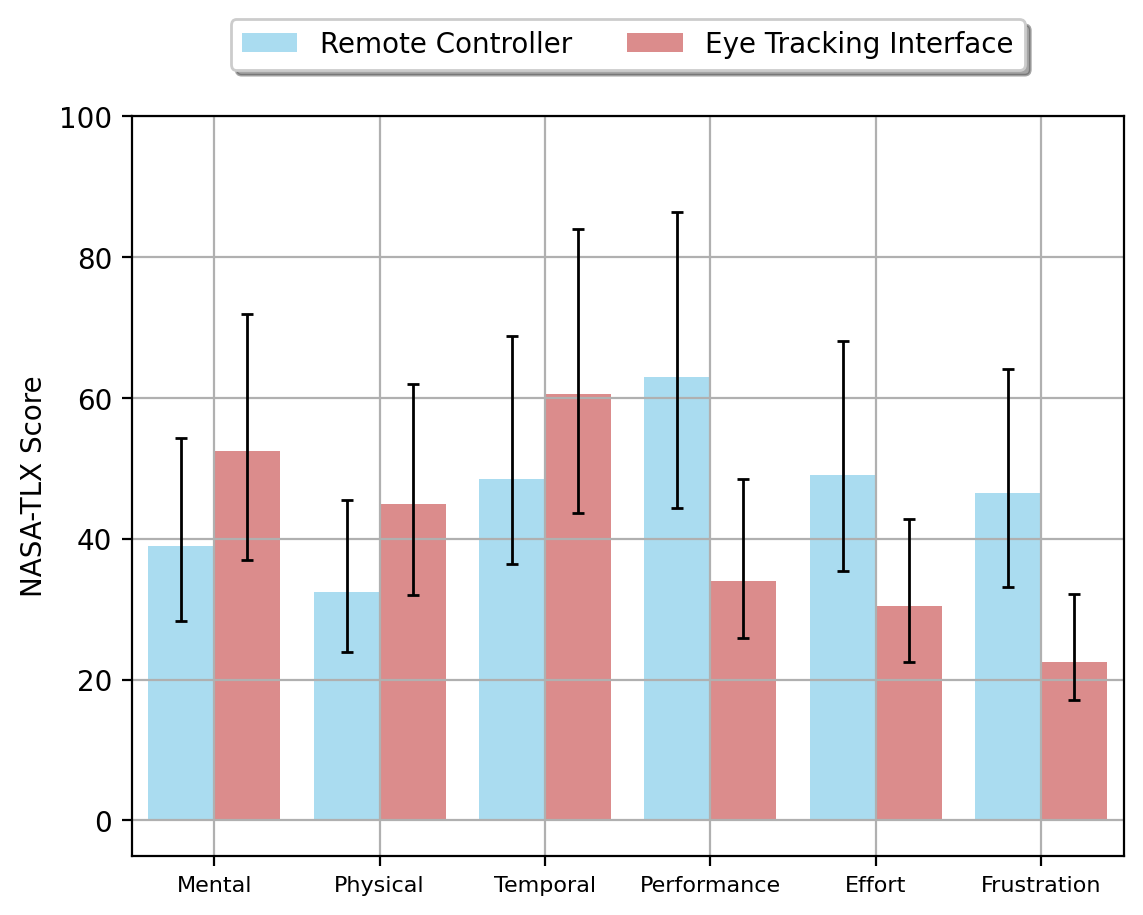}}
\caption{The visualization shows all dimensions of the NASA-TLX questionnaire. Black error bars denote a 95\% confidence interval (CI).}
\label{fig:NASA}
\end{figure}

\begin{figure}[t]
\centerline{\includegraphics[width=0.5\textwidth]{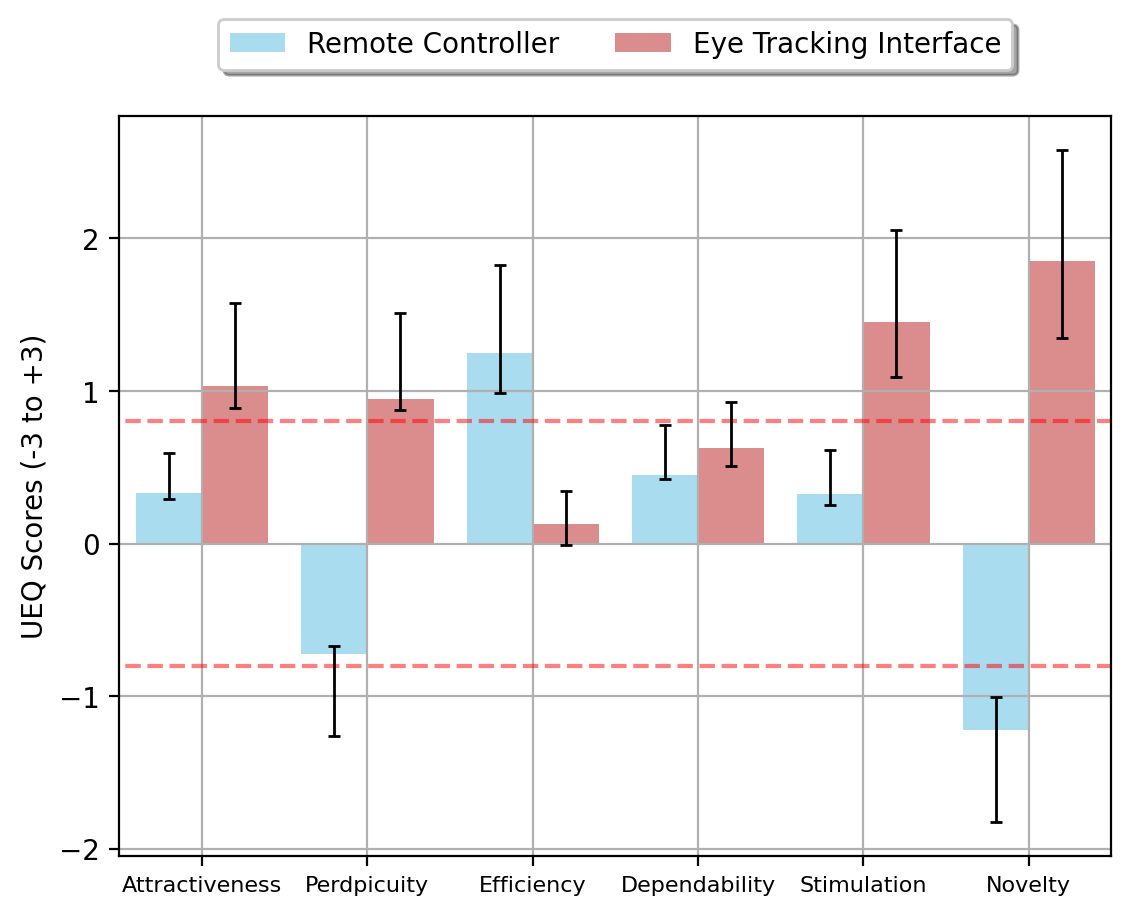}}
\caption{The visualization shows all dimensions of the UEQ questionnaire. Black error bars denote 95\% CI. The dotted red lines indicate UEQ's 0.8 threshold.}
\label{fig:UEQ}
\end{figure}

The NASA-TLX test results are presented in Fig. \ref{fig:NASA}. 
A Wilcoxon signed-rank test for paired samples on the NASA-TLX responses found significant effects on how participants perceived workload between the interfaces (V = 6.0, $p$ = $0.027 < 0.05$) resulting in a higher perceived workload in the case of the baseline remote controller (M = 46.4, SD = 6.7) compared to the eye-tracking interface (M = 40.8, SD = 4.3).
Users evaluated both physical (M = 45.0, SD = 8.1) and mental (M = 60.5, SD = 14.0) demand higher for the eye-tracking interface. Temporal demand was also perceived as higher for the developed system, however, statistical significance (V=10.0, $p$=0.083) was not demonstrated for this metric. On the other hand, the performance (V=0.0, $p$=0.007) and frustration (V = 2.5, $p$ = 0.02) were perceived by the responders as preferable when operating through the eye-tracking interface (M = 34.0, SD = 14.2) and (M = 30.5, SD = 9.2) compared to these respective metrics (M = 63.0, SD = 10.1 and M = 49.0, SD = 11.7) of the baseline remote controller. The UEQ dimensions for all drone control interfaces are visualized in Fig. \ref{fig:UEQ}. 

The UEQ evaluation shows that the eye-tracking received a positive user experience evaluation ($> 0.8$) for attractiveness, perspicuity, stimulation, and novelty. Most of the metrics for the remote controller including attractiveness, perspicuity, dependability, and stimulation were scored as neutral (in the range of -0.8 and 0.8), while the highest score was received for efficiency. The achieved result corresponds with the overall high performance of the experienced drone pilots with the controller. The acquired data did not show a statistical difference in dependability evaluation ($p$=0.69 $>$ 0.05). The results showed a significant difference and higher evaluation of the developed interface in the hedonic quality metric, while pragmatic quality did not show a significant difference, although evaluated higher for the eye-tracking interface. The possible reason for high evaluation could be that all groups of users were excited more with an unusual method of control and their accomplishment of controlling UAVs by vision rather than using a mediated interface.

\section{Conclusion and Future Work}

This research presents a novel approach named GazeRace to control race drones using a developed eye-tracking system based on real-time iris and eyelid-tracking frameworks. The system relies on a CV approach with a convolutional neural network to provide users with an easy-to-learn and immersive way to interact with racing drones. The results showed that the operators were able to effectively control the drone's motion, with the drone responding accurately to the user's gaze and eyelid motions. The developed system demonstrates the potential of eye-tracking technology to transform drone racing and other applications where intuitive, real-time control of drones is required. 

According to the experimental results, the eye-tracking interface was convenient for novice pilots. The results revealed that the users were able to decrease the length of the drone trajectory on average by 18\% (73.44 m from 89.29 m) while successfully navigating between the racing gates with the developed eye-tracking interface. It is also worth noting that, on average, there is not a significant difference between the use of remote control and using your eyes, around 3.5\%. This shows the equivalence of the two methods. While users evaluated both physical (M = 45.0, SD = 8.1) and mental (M = 60.5, SD = 14.0) demand higher for the eye-tracking interface, their performance and frustration were perceived as significantly more successful and less frustrating with this interface (M = 34.0, SD = 14.2) and (M = 30.5, SD = 9.2) compared to respective metrics (M = 63.0, SD = 10.1 and M = 49.0, SD = 11.7) with the baseline remote controller. 

In the next steps of our project, we plan to organize competitions in controlling a real drone, as well as attempt to enhance the drone's flight speed. There are limitations associated with the CV system, such as camera noise, frame rate, and image resolution. We will also address these issues to improve the control speed. The system's ability to interpret human gaze and eyelid motion could potentially be applied to other areas, such as augmented reality in robotics, drone navigation in complex environments, and new concepts of robot operation preserving the high mobility of the user.

\section*{Acknowledgements} 
Research reported in this publication was financially supported by the RSF grant No. 24-41-02039.








\balance{}

\bibliographystyle{IEEEtran}
\bibliography{reference} 

\end{document}